\documentclass[letterpaper, 10 pt, conference]{ieeeconf}  

\IEEEoverridecommandlockouts                              

\usepackage{graphics} 
\usepackage{epsfig} 
\usepackage{mathptmx} 
\usepackage{times} 
\usepackage{amsmath} 
\usepackage{amssymb}  
\usepackage{cite} 
\usepackage{mathtools} 
\usepackage{makecell}
\usepackage[hidelinks]{hyperref}
\usepackage[capitalise]{cleveref}
\usepackage{stmaryrd}

\crefname{equation}{}{}
\crefname{figure}{Fig.}{Figs.}

\usepackage{color} 
\usepackage{xcolor}
\definecolor{agreen}{rgb}{0.25, 0.51, 0.20}
\definecolor{better_blue}{rgb}{ 0.0627    0.5569    0.9098}

\title{\LARGE \bf
Optimizing Bipedal Maneuvers of Single Rigid-Body Models for Reinforcement Learning
}

\author{Ryan Batke, Fangzhou Yu, Jeremy Dao, Jonathan Hurst, Ross L. Hatton, Alan Fern,  and Kevin Green
\thanks{This work was supported in part by NSF Grants 1314109-DGE, 1849343-IIS, 1653220-CMMI, and 1826446-CMMI \newline
\- All authors are affiliated with Collaborative Robotics and Intelligent Systems Institute, Oregon State University, Corvallis, OR, USA. Email
        {\tt\small \{batker, greenkev, yufangzh, daoje, ross.hatton, alan.fern, jonathan.hurst\}@oregonstate.edu }}%
}

\begin{document}

\maketitle
\thispagestyle{empty}
\pagestyle{empty}


\begin{abstract}
In this work, we propose a method to generate reduced-order model reference trajectories for general classes of highly dynamic maneuvers for bipedal robots for use in sim-to-real reinforcement learning.
Our approach is to utilize a single rigid-body model (SRBM) to optimize libraries of trajectories offline to be used as expert references in the reward function of a learned policy.
This method translates the model's dynamically rich rotational and translational behaviour to a full-order robot model and successfully transfers to real hardware.
The SRBM's simplicity allows for fast iteration and refinement of behaviors, while the robustness of learning-based controllers allows for highly dynamic motions to be transferred to hardware.
Within this work we introduce a set of transferability constraints that amend the SRBM dynamics to actual bipedal robot hardware, our framework for creating optimal trajectories for a variety of highly dynamic maneuvers as well as our approach to integrating reference trajectories for a high-speed running reinforcement learning policy.
We validate our methods on the bipedal robot Cassie on which we were successfully able to demonstrate highly dynamic grounded running gaits up to 3.0 m/s.
\end{abstract}

\section{Introduction}
Bipedal animals from ratites to humans are capable of executing dynamic and aggressive motions that can seem effortless and graceful, but in reality require a complex balance of body momentum with fast and precise footstep placements.
Human athletes and animals are able to execute maneuvers where in a few steps they redirect their momentum in sharp, often successive turns to quickly change direction.
Additionally, both humans and many animals are capable of leaping into the air to reach higher locations, cross gaps or as a means to change direction.

\begin{figure}
    \centering
    \includegraphics[width=0.75\columnwidth]{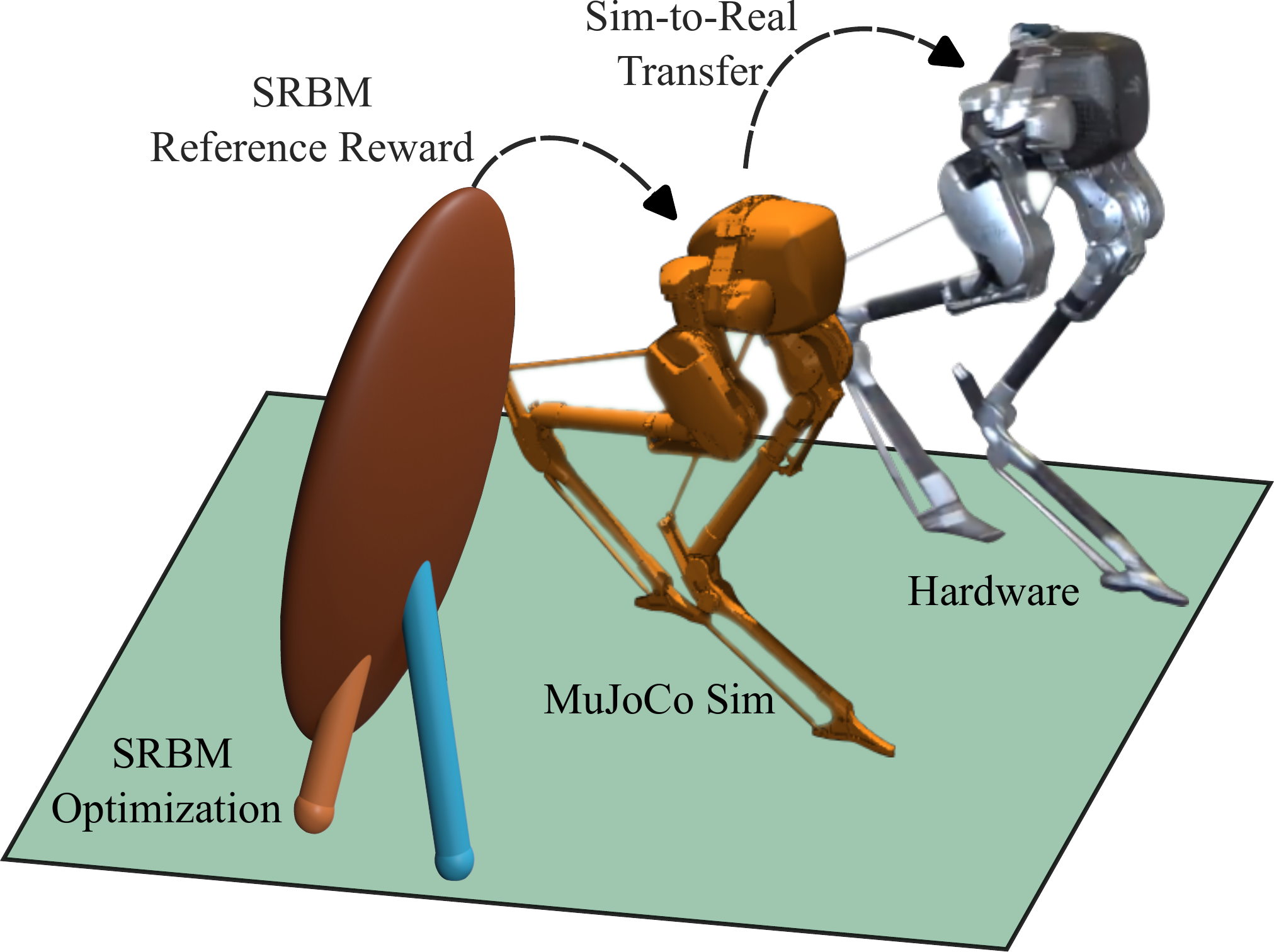}
    \caption{Dynamic maneuver workflow - SRBM optimized reference trajectories are used to develop learned controllers in simulation that are deployed on hardware. \vspace{-10 mm}}
    \label{fig:lead_image}
    \vspace{4mm}
\end{figure}

Bipedal robots have recently demonstrated increasingly dynamic capabilities \cite{siekmann2021sim, hwangbo2019learning}.
However, these advancements are still far behind the feats achieved by well tuned biological systems.
Modern control techniques that produce reliable locomotion such as model predictive controllers (MPC) based on Linear Inverted Pendulum (LIP) dynamics \cite{LIPMPC} severely restrict a systems behavior, preventing the realization of many high speed gaits, highly efficient gaits or dynamic maneuvers.

The usage of reduced-order-models (ROM) is a hallmark of optimization-based bipedal locomotion controllers.
These models serve to embed controllers with simplified descriptions of locomotion dynamics to reduce the complexity of planning.
Open-loop planning techniques like trajectory optimization (TO) can be used with a ROM to quickly generate highly non-linear optimal trajectories by enforcing meaningful constraints and objective functions on the underlying model.
The popular LIP model as well as the Spring Loaded Inverted Pendulum (SLIP model) lack the fidelity to capture the effects of angular momentum on the system that are necessary to achieve maneuvers such as sharp turns and spin jumps \cite{Geyer2006, kajita20013d}. 
Centroidal Momentum Models are another popular class of ROM commonly used for full humanoid robots \cite{MITDRCATLAS, WensingCMM}.
These models characterize both the linear and angular momentum of a robots CoM, but do not encode a mean orientation for the robot which is necessary for our target maneuvers. 

Data-driven methods offer a competing paradigm for control.
Recent successes with model-free deep reinforcement learning (DRL) based controllers for legged robots have demonstrated a wide range of robust and dynamic behaviours on hardware \cite{blind_stairs, ETH_hike, unsensed_loads}.
These learned controllers have the advantage of being trained in simulation on full-order robot models that contain dynamic information missing from ROMs.
Techniques such as dynamics randomization allow for a DRL agent to experience and adapt to various conditions over millions of iterations.
This allows for the training of incredibly robust policies that can reliably traverse the sim-to-real gap.
As powerful as these techniques have proven to be at demonstrating common gaits, the extension of reference-free DRL to structured maneuvers is still an unsolved problem. 

A promising alternative to relying solely on either ROM based optimizations or reference-free DRL is to utilize a dynamically rich model offline to first plan highly non-linear behaviors through TO.
These reference trajectories can then be used as a central part of a reward function to allow a learned controller to generalize the optimal trajectories from a simple model to the full-order robot and transfer to hardware.

In this work we present a single rigid-body model (SRBM) as applied to the bipedal robot Cassie. 
Using TO we create libraries of optimal trajectories offline that utilize the SRBM to plan a variety of highly dynamic behaviours such as running routes and jumps. 
These trajectories are then incorporated as part of a policies reward function and trained using proximal policy optimization (PPO) to develop controllers that are capable of executing the desired maneuver online on real hardware.
Finally, we test this process with the creation of a controller that achieves high speed dynamic running gaits.



\section{background}
\label{sec:background}
A SRBM approximates the inertia of a legged robot's many rigid bodies to a single body located at the center of mass (CoM) which is manipulated by ground reaction forces (GRFs) applied at foot locations through an idealized, massless leg.
The power of the SRBM is in its simplicity, using only a small number of parameters it is able to characterize the linear and rotational dynamics that are required to achieve highly aggressive maneuvers.
These models have been applied to quadruped robots to great success, allowing for hardware demonstrations of dynamic behaviours such as flips \cite{cheetah_flip}. 
SRBMs have been utilized for dynamic quadrupeds in conjunction with both optimization \cite{bledt2018cheetah} and RL based controllers \cite{xie2021glide}. 

Similar application of SRBMs are much less frequent in the domain of bipeds.
This is because, unlike quadrupeds, bipedal robot's legs often represent a more significant contribution to the robot's inertia \cite{tello_leg}.
A SRBM can still be used to plan for less-ideal bipeds and full humanoids, but requires careful consideration of how to best to apply the ROM to the full-order robot.
An impressive example of the SRBM being utilized to plan for angular momentum rich trajectories on a biped is Boston Dynamics' Atlas.
Atlas can perform a wide variety of maneuvers from backflips to sequential parkour jumps onto slanted surfaces \cite{ATLAS_parkour}.
Boston Dynamics has indicated that TO is used offline to create libraries of task-specific maneuvers based off of an SRBM that is adjusted for kinematic feasibility \cite{bdiATLAS, ICRA_ATLAS}.
Online the robot uses MPC guided by the offline optimal trajectories to perform short horizon optimizations that make real-time adjustments allowing for the execution of the dynamic behaviours on hardware. 

Previous work combined reference trajectories and DRL to develop walking controllers for Cassie \cite{UBC_collab}, but these were based off only a single manipulated reference and thus were limited in application.
Additional work was conducted using SLIP models to create libraries of reference trajectories to guide DRL \cite{Learned_SLIP}.
Most similar to this current work, combining TO and DRL for locomotion was shown by \cite{TO+RL} for a terrain-aware quadrupedal robot.

\section{Single Rigid Body Model Formulation}
\label{sec:centroidalModel}
Our representation, dynamics and quaternion integration method are inspired by \cite{Jackson2021PlanningWithAttitude}.
This body has a configuration space of $SE(3)$ and a velocity in $\mathbb{R}^6$.
While this is the precise structure of the space, we elect to use a common representation of the configuration space as $[\mathbf{p_c}, \mathbf{q}] \in \mathbb{R}{^3} \times \mathbb{S}{^3}$.
Here $\mathbf{p_c}$ is the 3D position of the body's center of mass (CoM) and $\mathbf{q}$ is the body's attitude as a quaternion.
The body's velocity is represented by the CoM linear velocity $\mathbf{v_c} \in \mathbb{R}{^3}$ and the body frame angular velocity  $\mathbf{\omega} \in \mathbb{R}{^3}$.
We then combine the configuration and velocity into a single state vector $\mathbf{x}$ which has dynamics
\begin{equation}
    \begin{aligned}
        \mathbf{x} = \begin{bmatrix}
            \mathbf{p_c}\\
            \mathbf{q}\\
            \mathbf{v_c}\\
            \mathbf{\omega}
        \end{bmatrix}, && \mathbf{\dot{x}} = \begin{bmatrix}
            \mathbf{v_c}\\
            \frac{1}{2}\textbf{q} \otimes \hat{\omega}\\\
            \frac{1}{m}\mathbf{F}_W(\mathbf{x},\mathbf{u}) \\
            J^{-1}(\mathbf{\tau}_B(\mathbf{x},\mathbf{u}) - \mathbf{\omega} \times J\mathbf{\omega})
        \end{bmatrix}.
    \end{aligned}
    \label{eq:centroidal_state}
\end{equation}
Here $\mathbf{x}$ and $\mathbf{u}$ are the state and control vectors, $m$ is the mass, $\mathbf{J} \in \mathbb{R}{^{3 \times 3}}$ is the body frame rotational inertia matrix, $\mathbf{F}_W(\mathbf{x},\mathbf{u}) \in \mathbb{R}{^3}$ are the external forces in the world frame, and $\mathbf{\tau}_B(\mathbf{x},\mathbf{u})$ are the external moments as expressed in the body frame.
In the rotational dynamics, $\otimes$ represents quaternion multiplication and $\hat{\omega}$ is the angular velocity as a quaternion with zero scalar part.
More details on the rotational dynamics can be found in \cite{Jackson2021PlanningWithAttitude}.

Our model is actuated by ground reaction forces which are applied by a massless, ideal leg.
These ground reaction forces can be thought of as a perfect force vector applied to the body through the foot's contact point.
This can then be transformed into a wrench at the body to integrate the dynamics. 
In our implementation the model can have one, two or no active foot contacts depending on its specified hybrid mode.

\section{Trajectory Optimization Formulation}
\label{sec:trajOpt} 
We used direct collocation trajectory optimization implemented in COALSECE \cite{coalesce} and solved using IPOPT \cite{IPOPT} to create libraries of dynamically feasible open-loop trajectories offline. 
Dynamically rich trajectories can be generated though specifying a physically meaningful objective function, a sequence of hybrid modes and constraints that shape the behavior of the model to the dynamic maneuvers that are the goal of this work. 
By specifying sequences of hybrid phases and applying sets of constraints, through the optimization we are able to develop a wide variety of agile maneuvers that can be easily iterated and shaped to create expert references for learning-based controllers.



\subsection{Hybrid Modes} 

We are interested in agile maneuvers that span multiple footsteps which means that our motions will span multiple hybrid modes.
While there is much interest in contact implicit TO, we chose to prescribe the contact sequence a priori.
We chose this because the space of contact sequences for point foot bipedal models with no arm contact in a flat environment is relatively small.
For standard locomotion gaits at a commanded speed there are three reasonable options:
The robot can walk with a double stance phase, it can run with an aerial phase, or it can perform a grounded run with no double stance or flight phase.
Each of these corresponds to a different sequence of three hybrid modes: aerial flight, single stance, double stance. 
Our model's massless, ideal legs mean that we can consider left and right single stances to have identical dynamics.
In this work we are interested in agile maneuvers, as well as dynamic jumps. 
For running and turning maneuvers we use a grounded run contact pattern, which consists of alternating single stance modes.
For jumping maneuvers we use double stance and aerial flight modes.

\subsection{Decision Variables}
Our TO consists of several hybrid modes serially appended to each other, with a total of $M$ modes for any maneuver.
Each of these modes has $N$ collocation points which evenly divide the duration of the hybrid mode in question.
This means that for any mode we have $N$ sequential variants of the state and inputs.
Our state is parameterized by $13$ components: $3$ from CoM position ($\mathbf{p}_{c}$), $3$ from CoM velocity ($\mathbf{v}_{c}$), $4$ from the orientation quaternion ($\mathbf{q}$), and $3$ from body frame angular velocity ($\mathbf{\omega}$). 
If we are in single-stance we need to collocate the inputs, which in this case are $N$ of the ground reaction forces $F \in \mathbb{R}^3$.
In order to adjust the timing of the hybrid transitions the optimizer needs to be able to adjust the duration of the phases. 
To this end, we also create a single decision variable per phase which represents the phase's duration.
Lastly, the optimization needs to be able to adjust the location where the stance foot is placed.
This is accomplished by adding a single position vector as a decision variable for single-stance phases $\mathbf{p}_f \in \mathbb{R}^3$.
For double stance phases we have two foot positions, $\mathbf{p}_{f,1}, \; \mathbf{p}_{f,2}$, and similarly $F_1 , \; F_2$.
In this work we constrain the vertical components of foot position to be zero for flat ground gaits.
This results in a total decision variable count of $13N + 1$ for flight phases, $16N + 4$ for single-stance phases, and $19N + 7$ for double stance.

The indexing $x(m,n)$ is used in this paper to refer to the $m^{th}$ hybrid mode and the $n^{th}$ collocation point of any variable.
When no indexing is used this refers to a variable throughout the entire trajectory, for each mode and collocation point.
The letters $i$ and $F$ refer to the initial and final modes of any maneuver, where $1$, and $N$ are used to indicate the first and final collocation point of any mode. 

\begin{figure*}
    \centering
  \includegraphics[width=\textwidth]{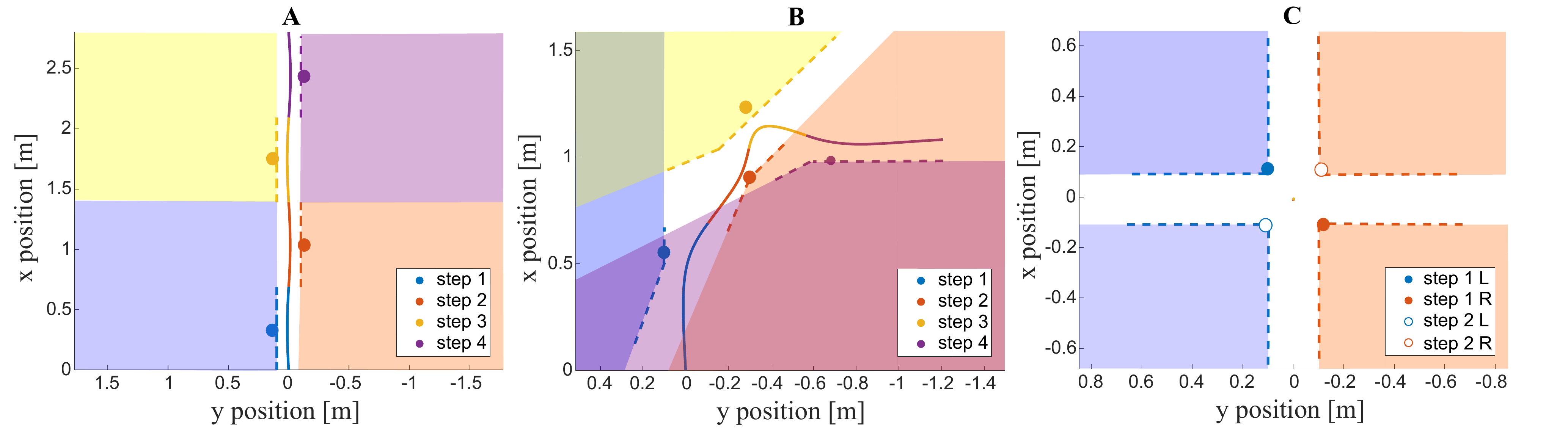}
  \caption{Footstep constraint visualizations, A: forward running, B: $90^{0}$ turn, C: $90^{0}$ spin jump. These constraints are built of a pair of linear constraints on foot location in X-Y space. The constraints are based on a ``nominal'' heading and are relative to the initial and final positions of the body in that dynamic phase.}
    \label{fig:footstep_constraints}
    \vspace{-5mm}
\end{figure*}

\subsection{Transferability Constraints}
\label{sec:transfer} 
In order to constrain the model to produce maneuvers which are more likely to transfer to the physical robot we apply some general constraints. 
First is kinematic feasibility:
The workspace of most real robot legs are complicated spaces, particularly so for Cassie's four bar leg mechanism.
We would also need to describe this workspace relative to the center of mass of the robot.
Knowing the difficulty in creating a very accurate model, we compromise and impose a single constraint on the total leg length,
\begin{equation}
    \Vert \mathbf{p}_c - \mathbf{p}_f \Vert \leq L_\text{max}
    \label{eq:centroidal_legLength}.
\end{equation}

A constraint is also imposed on the angle that the leg can make with respect to the unit vector pointing directly downwards in the body frame, $\mathbf{\hat{z}_{body}}$.
This is used as a proxy for joint and configuration limits.
A variable $\psi$ is used to define the bounds of how different the normalized leg vector can be from $\mathbf{\hat{z}_{body}}$, 
\begin{equation}
    \psi < \mathbf{\hat{z}}_{\text{body}} \cdot \frac{(\mathbf{p}_c - \mathbf{p}_f)}{\Vert (\mathbf{p}_c - \mathbf{p}_f) \Vert}.
\end{equation}

Next we impose several constraints on the ground reaction forces.
We apply a quadratic friction cone constraint,
\begin{equation}
    \mu F_z^2 \geq F_x^2 + F_y^2.
    \label{eq:centroidal_friction_cone}
\end{equation}
Additionally, we apply a maximum force constraint based on the physical limitations of our actuation
\begin{equation}
    F_z^2 + F_x^2 + F_y^2 \leq F_\text{max}^2.
    \label{eq:max_leg_force}
\end{equation}

Our simplified model has no actuator dynamics which means it is able to instantly change the foot force which is not realistic.
We impose a maximum yank constraint (rate of change on vertical force) $\dot{F}_{max}$ based on analysis of our previously developed controllers for the Cassie robot.
\begin{equation}
    - \dot{F}_{\text{max}} \leq \dot{F}_z \leq \dot{F}_{\text{max}}
    \label{eq:centroidal_force_rate}
\end{equation}
In reality we impose the constraint on the finite difference between subsequent collocated inputs, because the force has no intrinsic dynamics.
Feasibility constraints are placed on each component of $\mathbf{\omega}$ to prevent the optimizer from exploiting the simplified models dynamics, this also yields smoother more feasible motions. We apply a simple bound on the angular velocity, $\lvert \mathbf{\omega} \rvert \leq \omega_{\text{max}}$.

Lastly, a simple footstep heuristic is used to present the optimization with areas of feasible space to place the models feet.
This constraint ensures that feet are placed on the correct side of the body to prevent leg crossing. 
The heuristic defines acceptable regions by first constructing two half-planes for each step, drawn from the CoM position at foot touchdown and foot liftoff and extended out infinitely along the current heading.
These planes are then projected out in the correct side of the body by some minimum distance $\delta_{\text{min}}$.
The region encapsulated by the intersection of these planes forms a feasible region where feet may be placed.
$\delta_{\text{min}}$ is needed to prevent the optimizer from exploiting monopod dynamics and placing footstep locations directly under the model which does not translate well to actual biped hardware. \cref{fig:footstep_constraints} depicts a visualization of these constraints for several example maneuvers.

\subsection{Composing Maneuvers}
We are able to design agile maneuvers within our TO framework by specifying sequences of hybrid modes and accompanying sets of constraints that shape the optimization to produce the desired behavior.
The transferability constraints from \cref{sec:transfer} are applied in general to each hybrid mode. 
Additionally, loosely bounded constraints are also placed on the minimum and maximum duration for each phase. 
These durations control the stepping frequencies for single-stance phases, and the liftoff, flight, and touchdown durations for jumping motions.
Any number of maneuvers can be developed using these principles.
For breadth we present three formulations for hybrid mode sequences and constraint sets to create trajectories for nominal forward running gaits, dynamic turns, and spin jumps.

\subsubsection{Forward Running}
To develop a grounded running gait we specify sequences of single-stance dynamics and apply the appropriate set of constraints.
Constraints are enforced to ensure that the heading remains mostly constant throughout the gait. 
To enforce constraints on quaternion orientations, we employ a distance function which bounds the angle difference $\theta_{\text{tol}}$ that two quaternions can have from each other,
\begin{equation}
    d( \mathbf{q}_{1}, \mathbf{q}_{2}) \leq \theta_{\text{tol}}. 
\end{equation}
We use this function to formulate a constraint between the initial orientation $\mathbf{q}(i,1)$ and a specified desired orientation for forward running $\mathbf{q}_{\text{run}}$, as well as the final orientation $\mathbf{q}(F,N)$ and  the same desired orientation.
Average velocity constraints are applied in the form of,
\begin{equation}
    \frac{p_x(F,N) - {p_x(i,1)}}{T} = v_{\text{des}},
\end{equation}
where T is the total duration of the maneuver.
Additionally, zero average velocity constraints are applied in the lateral direction to $p_y$.
Cyclic equality constraints are applied within each phase and enforced on the bodies height, $p_{z}(m,1) = p_{z}(m,N)$ and vertical velocity, $v_{z}(m,1) = v_{z}(m,N)$. 
Similar constraints are applied to the angular velocity in the $y$ direction, while the $x$ and $z$ angular velocities are mirrored with constraints such as $\omega_x(m,N) = -\omega_x(m,1)$

For each hybrid mode the body's final orientation $\mathbf{q}(m,N)$ is mirrored about it's sagittal plane.
Using the distance function $d$, this mirror, $\mathbf{q}_{\text{mirror}}(m,N)$ must be within a tolerance $\theta_{\text{mirror}}$ of the initial orientation of the mode $\mathbf{q}(m,1)$,
\begin{equation}
    d( \mathbf{q}_{\text{mirror}}(m,N), \mathbf{q}(m,1)) \leq \theta_{\text{mirror}}.
\end{equation}

\subsubsection{Turning}
Turning maneuvers are sequences of single-stance phases with the explicit desire to change heading throughout the maneuver.
The desired heading for each mode is updated by incrementing the current heading by the desired total change in heading over the number of steps for the turn.
Using our quaternion distance function $d$ we ensure that 
that the bodies orientation at the end of the maneuver $\mathbf{q}(F,N)$ aligns with the desired orientation for the turn $\mathbf{q}_{\text{turn}}$ within a tolerance $\theta_{\text{turn}}$.
Desired velocity equality constraints are only enforced on the initial and final collocation points, $\mathbf{v}(i,1)$ and $\mathbf{v}(F,N)$, as the optimization needs to be free to modulate the bodies velocity throughout the maneuver.
Cyclic equality constraints are only placed on the initial and final heights for the maneuver
\begin{equation}
    p_{z}(i,1) = p_{z}(F,N).
\end{equation}

\subsubsection{Spin Jumps}
Our example formulation for jumping maneuvers consist of $4$ hybrid modes, a double stance for liftoff, followed by two consecutive flight phases, and ending with a double stance for touchdown.
Two flight phases are required so that the jump height can be precisely specified at apex.

Multiple constraints are enforced on the models orientation at different stages of the jump.
Our distance function $d$ is used on the final orientation at liftoff, $\mathbf{q}(i,N)$, to constrain it to be within some tolerance $\theta_{\text{liftoff}}$ of a desired neutral starting position $\mathbf{q}_{\text{liftoff}}$.
A loose tolerance $\theta_{\text{touchdown}_{i}}$ is applied using the same quaternion distance constraint at the beginning of touchdown $\mathbf{q}(F,1)$ to ensure that the majority of the turn is accomplished in the air to achieve the desired rotation $\mathbf{q}_{\text{touchdown}}$.
A second tighter tolerance $\theta_{\text{touchdown}_{F}}$ applied to the final orientation $\mathbf{q}(F,N)$ ensures that the model makes any required corrections to finish close to $\mathbf{q}_{\text{touchdown}}$.

Equality constraints are placed on $v_{c}(i,1), v_{c}(F,N), \omega(i,1),$ and $\omega(F,N)$ to enforce that the jumps start  and end at rest.
Lastly, equality constraints are placed at the final collocation point of the first aerial phase (hybrid mode 2). 
These ensure that $p_{z}(2,N)$ reaches the desired height, $p_{\text{z}_{\text{des}}}$, and is at the same time at the apex of the jump by forcing $v_{z}(2,N)$ to $0$.
Additional constraints can also be employed to ensure displacement of the body in any direction, as would be required to jump up onto a surface or over a gap.

\subsection{Objective}
\label{sec:obj}
The objective for each trajectory is the minimization of the squared ground reaction forces and the resulting moments applied at the body. 
This objective incentivises smoother control inputs which are more transferable to hardware. 
The objective function is the integral
\begin{equation}
    f = \int_{0}^{T} \big( u^{2}(\tau) + M^{2}(\tau) \big) d\tau,
\end{equation}
which is approximated using trapezoidal integration. 

\subsection{Library Generation}
The end goal of the optimization is to construct entire libraries of optimal trajectories for a given maneuver. 
For the case of grounded maneuvers such as the running gaits and dynamic turns, these libraries are similar optimizations sweeping through a range of desired velocities.
For the spin jumps the optimization could be solved for over a range of desired heights or end displacements of the CoM.
The purpose of generating entire libraries for maneuvers is to make them more generalizable and help develop more robust learned control policies that can execute the target maneuver on hardware and for a wide variety of conditions.

\begin{figure}
    \centering
    \vspace{4 pt}
    \includegraphics[width=0.85\columnwidth]{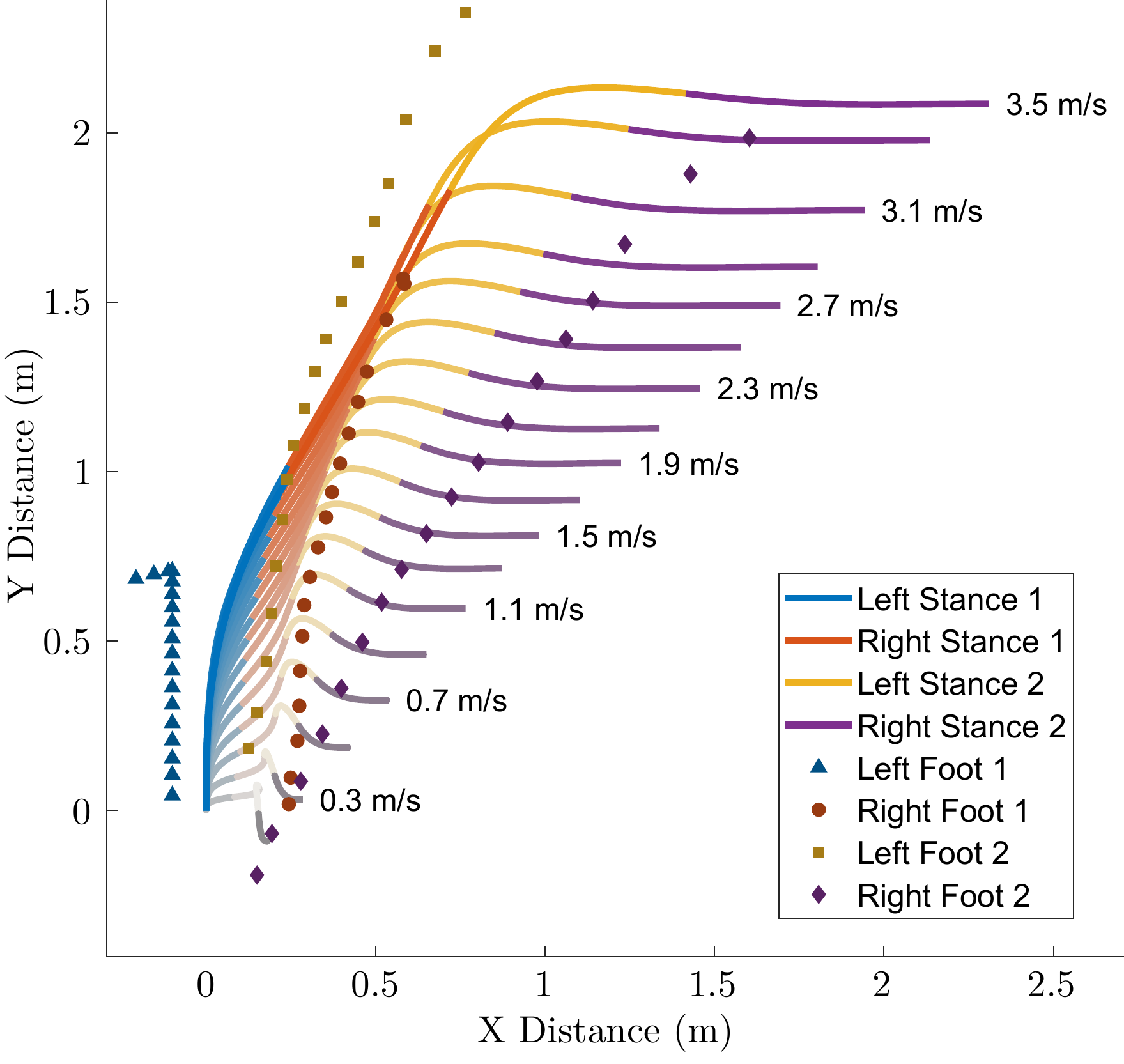}
    \caption{Center of mass traces and footstep locations for a library of 4-step $90^{\circ}$ turning trajectories.}
    \label{fig:turning_library}
\end{figure}

Much of the success of solving a TO problem lies with good choice of initial conditions. 
Initial guesses are expertly chosen for the first optimization of a library.
Then the solution is used as the initial guess for the next iteration to speed-up convergence.
This additionally increases the likelihood that the library of trajectories will smoothly vary which is useful for the DRL problem.
\cref{fig:turning_library} is an example of trajectories for a library of 4-step $90^{\circ}$ turns from 0.1-3.5 m/s.

\begin{figure*}
    \centering
  \includegraphics[width=\textwidth]{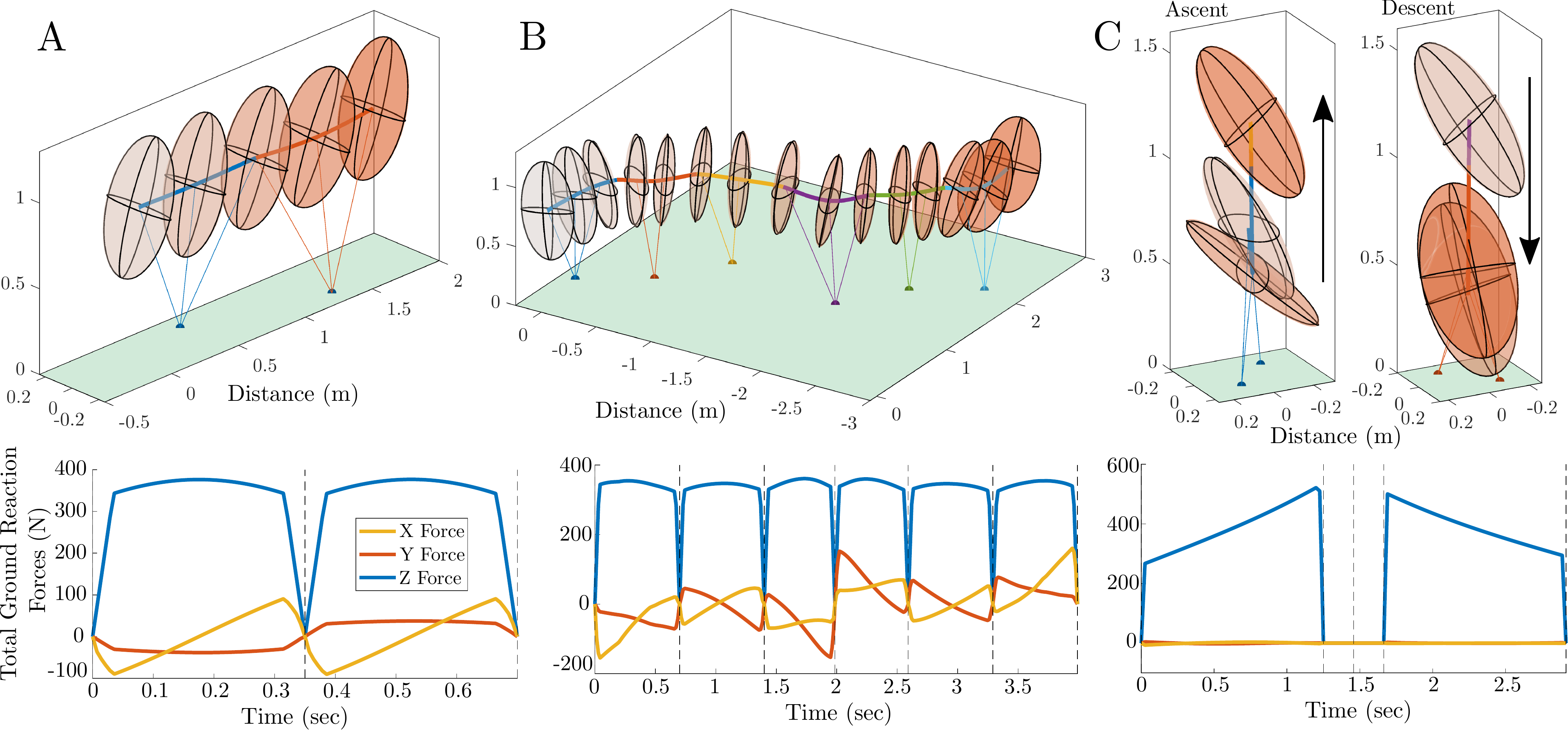}
  \caption{Visualizations and accompanying GRF plots for a selection of maneuvers. A: 2-step running trajectory at 2 m/s. B: 3-step $-90^{\circ}$ turn into 3-step $+90^{\circ}$ turn at 2 m/s. C: $-90^{\circ}$ spin jump, CoM reaching 1.2 m. Vertical dashed lines on GRF plots indicate the transition between hybrid phases.}
    \label{fig:maneuver_viz}
    \vspace{2 mm}
\end{figure*}

\section{Reinforcement Learning Problem}
\label{sec:RL}

Our approach to developing reference guided RL control policies borrows ideas from both referenced-based work \cite{UBC_collab} as well as more recent reference-free work \cite{blind_stairs}.
We seek to leverage the rich dynamic information from the SRBM captured in maneuver libraries to create meaningful reward functions that not only bootstrap the learning process from this expert information but are also capable of extending the capabilities of learned controllers to highly dynamic maneuvers.
In this work, as a case study to validate the transferability of SRBM dynamics to Cassie we attempt the task of developing a high-speed running policy from a library of optimized running references from 0.1-3.5 m/s.
Although running can be described entirely as a periodic phenomenon without ROM information \cite{siekmann2021sim}, we first seek to learn how well SRBM reference policies cross the sim-to-real gap, and if utilizing this reference information provides any benefits to training over reference-free methods.


\subsection{Problem formulation}
\label{sec:learning problem}
Our control policy is a long-short-term-memory (LSTM) NN with two fully connected hidden layers of 128 units each.
The policy takes as input both estimated state and task information. 
$41$ total inputs represent the robot's internal state, comprised of; pelvis orientation, pelvis rotational velocity, estimated foot positions, motor positions and velocities and actuated joint positions and velocities. 
Three additional inputs provide task information in the form of a commanded velocity and a 2D clock signal. 
The outputs of the network are the $10$ commanded joint PD set points which are sent to high rate, low-level motor controllers.

Control policies are trained using a MuJoCo simulated model of Cassie\footnote{Simulation available at \href{https://github.com/osudrl/cassie-mujoco-sim}{https://github.com/osudrl/cassie-mujoco-sim}}. 
The learned controller runs at 40 Hz, with a total of 250 steps per episode (equivalent of 6.25 sec).
The reference trajectories range in length from 0.54-1.0 sec and are looped  repeatedly for the duration of each episode.

At the beginning of each episode Cassie is initialized with a random state from a previously trained nominal walking gait.
This randomization helps to increase the robustness of the policy as it most often needs to recover from poor starting conditions at the beginning of each episode.
Additionally the policy is given a random commanded velocity which selects the corresponding reference trajectory.
 
\subsection{Learning Procedure}
\label{sec:learning procedure}

We use proximal policy optimization (PPO) \cite{PPO} with the Adam optimizer due to its relative simplicity and previously demonstrated successes. 
Training hyperparameters provided in  \cref{table:hyperparameters}.

\begin{table}
\centering
\begin{tabular}{|l c|} 
 \hline
 Parameter & Value \\ [0.5ex] 
 \hline
 Adam discount ($\gamma$) & 0.95 \\ 

 Adam epsilon & 1 $\times$ $10^{-6}$ \\
 
 actor learning rate & 3 $\times$ $10^{-4}$ \\
 
 critic learning rate & 3 $\times$ $10^{-4}$ \\
 
 gradient update clipping & 0.05 \\
 
 batch size & 64 \\
 
 epochs & 5 \\ 
 
 sample size & 50000 \\
 \hline
\end{tabular} 
\caption{PPO hyperparameters}
\label{table:hyperparameters}
\vspace{-5mm}
\end{table}

We formulate our reward function to include components that reward tracking the important dynamics from the reference trajectories.
We include terms for matching body orientation \cref{eq:reward_quat}, CoM linear velocities \cref{eq:reward_linvel} and angular momentum $\mathbf{L} = J\mathbf{\omega}$ \cref{eq:reward_angMom}, as well as $x$ and $y$ components of the foot positions relative to the body \cref{eq:reward_foot}.
\begin{equation} 
    r_{\mathbf{q}} = 0.05 \, \exp(-d(\mathbf{q}, \mathbf{q}^{\text{ref}}) ) 
    \label{eq:reward_quat}
\end{equation}
\begin{align}
    \begin{split}
    r_{v} = &\; 0.35 \, \exp(-\lvert v_{x} - v_{x}^{\text{ref}}\rvert) \\
    &+ 0.1 \, \exp(-\lvert v_{y} - v_{y}^{\text{ref}}\rvert) \\
    &+ 0.1 \, \exp(-\lvert v_{z} - v_{z}^{\text{ref}}\rvert)
    \end{split}
    \label{eq:reward_linvel}
\end{align}

\begin{equation}
    r_{\mathbf{L}} = 0.15 \, \exp(-\lVert \mathbf{L} - \mathbf{L}^{\text{ref}}\rVert)  
    \label{eq:reward_angMom}
\end{equation}
\begin{align}
    \begin{split}
    r_{p_{f}} = 0.15 \, \exp(-\lvert 20 \cdot (p_{f_{x}} - p_{f_{x}}^{\text{ref}})\rvert) \\
    + 0.15 \, \exp(-\lvert 20 \cdot (p_{f_{y}} - p_{f_{y}}^{\text{ref}})\rvert).
    \end{split}
    \label{eq:reward_foot}
\end{align}

\begin{figure*}
    \centering
  \includegraphics[width=0.7\textwidth]{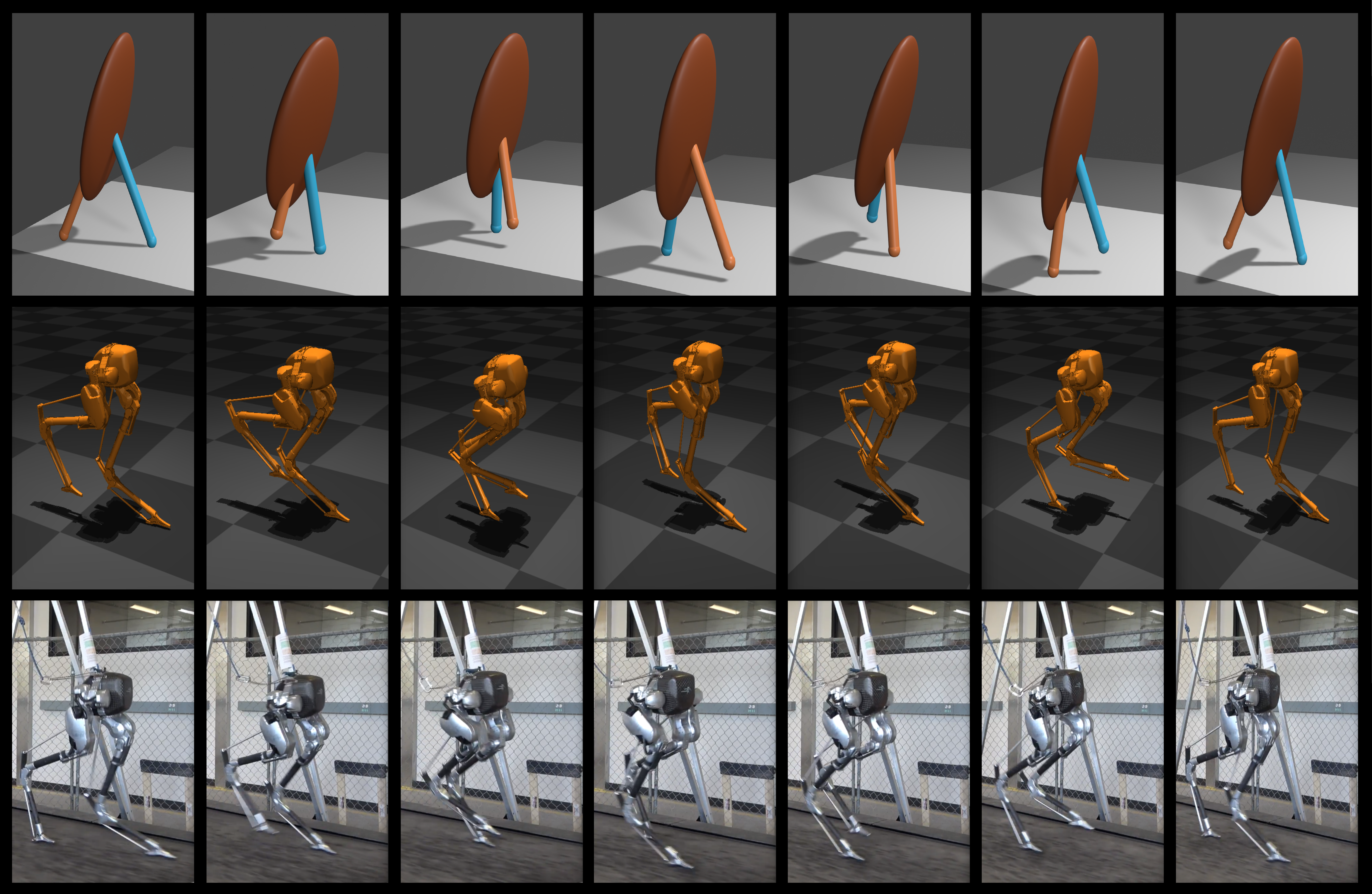}
  \caption{Comparison of 3.0 m/s running. Top: Render of single rigid-body model optimized trajectory. Middle: Cassie MuJoCo simulation. Bottom: Cassie hardware treadmill test.}
  \label{figure:results_film_strip}
\end{figure*}

Additionally our controller features a piecewise-linear clock signal explained in detail in \cite{unsensed_loads} that rewards the agent based on matching the stepping frequencies determined by the optimization.
\begin{equation}
    r_{\text{clock}} = 0.3 \, \exp(- \lvert F_{\text{clock}} \rvert ).
\end{equation}

Non-reference terms are also added to account for limitations inherent to model itself as well as the limitations of its application to Cassie.
Firstly, the model only has point feet and no swing leg dynamics, to account for these we reward foot orientation and swing apex heights to encourage the system to keep its feet facing forward and for the robot to not shuffle its feet.
\begin{equation} 
    r_{\mathbf{q}} = 0.3 \, \exp(-\lvert d(\mathbf{q}_{\text{foot}}, \mathbf{q}_{\text{foot}}^{\text{ref}})\rvert ) 
\end{equation}
\begin{equation}
    r_{z_{\text{foot}}} = 0.3 \, \exp(- \lvert z_{\text{foot}} - z_{\text{foot}}^{\text{des}} \rvert).
\end{equation}

A major point of consideration is how to impart single rigid-body orientations onto a multiple-rigid body complex system.
Geometric mechanics techniques such as minimum perturbation coordinates \cite{Travers2013} possibly present intriguing insights but are out of scope for this work.
To this effect we chose to match our orientations to Cassie's pelvis, which only represents a portion of the robots total orientation, but is the single most natural choice of its many rigid bodies.
As the pelvis is not the optimal choice to apply the SRBM rotational information we found it necessary to include terms to our reward formulation that attempt to ameliorate this discrepancy. 
A lateral drift term is included to keep the robot running in a straight line,
 \begin{equation}
    r_{\text{drift}} = \begin{cases}
             0.3  & \text{if } \lvert p_{y} \rvert < 0.2 \\
             0.3 \, \exp(- \lvert 15 \cdot p_{y} \rvert )  & \text{if }\lvert p_{y} \rvert \geq 0.2 \;.
       \end{cases}
\end{equation}
The final rewards are penalties to hip roll motor velocities to prevent excessive motions at the pelvis.
\begin{equation}
    r_{\text{hip roll}} = 0.1 \, \exp(- \lvert \omega_{\text{hip roll}} \rvert )
\end{equation}
\begin{equation}
    r_{\text{hip yaw}} = 0.1 \, \exp(-\lvert \omega_{\text{hip yaw}}\rvert).
\end{equation}

All reward terms are then normalized before summing to calculate the total reward.
In order to increase our policies robustness and aid in the crossing of the sim-to-real gap we employ dynamics randomization during training.
At each episode the dynamics listed in \cref{Tab:dynamics_rand} are randomly modified to aid in the generalization of our policy and its ability to overcome discrepancies from simulation to hardware:

\begin{table}
\centering
\begin{tabular}{|l  c|} 
 \hline
 Parameter & Randomization Bounds \\ [0.5ex] 
 \hline
 Policy Rate & [0.95:1.05] $\times$ default \\

 Joint Damping & [0.5:3.5] $\times$ default \\

 Joint Mass & [0.5:1.5] $\times$ default \\

 Ground Friction & [0.35:1.1] \\ 
 \hline
\end{tabular}
\caption{Dynamics randomization parameter ranges.}
\label{Tab:dynamics_rand}
\vspace{-5mm}
\end{table}

\section{Results}
\label{sec:results}
Using our dynamic maneuver TO formulation outlined in \cref{sec:trajOpt} we are able to create a wide variety of reference maneuvers based on SRBM dynamics as applied to the Cassie robot.
Several such example maneuvers and their ground reaction forces are presented in  \cref{fig:maneuver_viz} and in the supplemental video.

\subsection{Simulation Results}
\label{sec:results_sim}
\begin{figure}
    \centering
    \vspace{4 pt}
    \includegraphics[width=1.0\columnwidth]{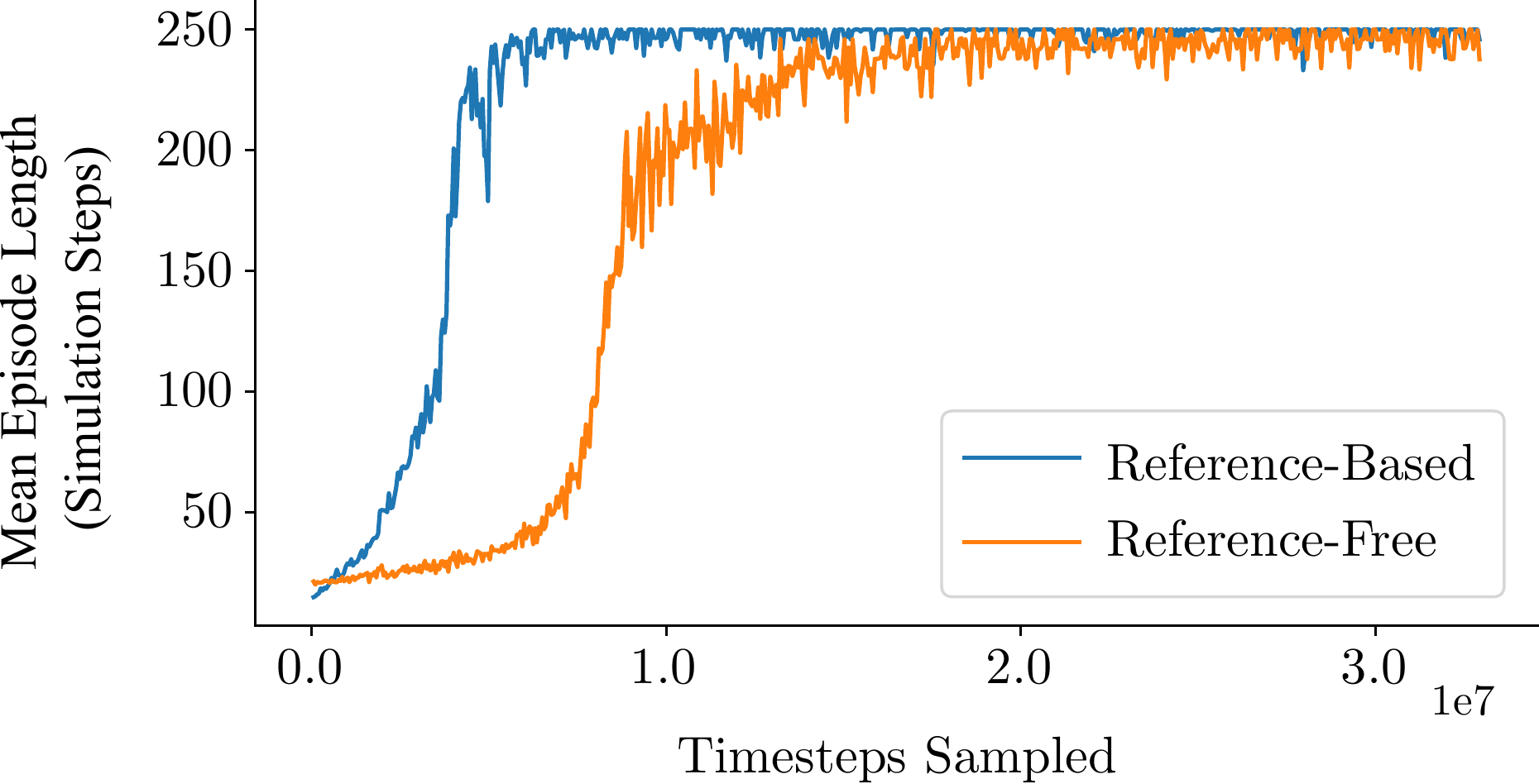}
    \caption{Mean training episode length for grounded running trained on speeds between 0.1 and 3.5 m/s. The episode length is an useful way to compare ability of controllers with different reward functions.}
    \vspace{-5 mm}
    \label{fig:training_time}
\end{figure}

Two running policies were trained with the same inputs and hyperparameters over the velocity range of 0.1-3.5 m/s. 
First is our reference-based control policy with the same composition as outlined in \cref{sec:RL}.
The second serves as a reference-free baseline and utilized the previous state of the art reward composition laid out in \cite{siekmann2021sim}.
Given the differences in reward terms, we elect to compare mean episode length for each policy during training as a metric for quantifying the stability of each policies gait, as shown in  \cref{fig:training_time}.
We can note that the reference-based policy achieves a stable running gait (consistently reaches the maximum episode length of 250 simulation steps) with much greater sample efficiency than the reference-free policy.
A stable running gait is achieved by the reference-based policy after only $\sim0.8\times10^{7}$ timesteps where the reference-free policy required $\sim2.0\times10^{7}$ timesteps, indicating that for this particular gait the usage of SRBM reference dynamics yielded a roughly $2.5\times$ increase in sample efficiency.
From qualitatively analyzing both policies in simulation, we note that the reference-based policy moves in a more fluid manner, characterized by the natural oscillating motions of the model when compared to the more rigid reference-free policy. 
This can be seen in the supplemental video.

\subsection{Hardware Results}
\label{sec:results_hardware}
The reference-based control policy was successfully deployed onto a physical Cassie robot and was able to run on a treadmill up to 3.0 m/s as shown in the supplemental video.
A high level of similarity can be visually seen between the gaits of the SRBM reference (with added swing leg kinematics for visualization), the MuJoCo simulation in which training took place, and the physical hardware test, as shown in \cref{figure:results_film_strip}. 
This demonstration indicates that SRBM dynamic reference information, even with rotational dynamics applied only at Cassie's pelvis can indeed be used in a RL framework to train dynamic running policies up to high speeds.
However, we were not able to fully cross the sim-to-real gap as top speeds (between 3.0 and 3.5 m/s) achieved in simulation were not able to be repeated on hardware. 

\section{Conclusion}
\label{sec:conclusion}


In this work we presented a framework for both authoring and executing dynamic maneuvers for bipedal robots.
Using a single rigid-body model-based trajectory optimization amended for bipeds, we formulated a method of developing reference trajectories for arbitrary maneuvers by specifying sequences of hybrid modes with sets of shaping constraints.
We were also able to demonstrate that these reference trajectories could be used to develop reinforcement learning control policies capable of crossing the sim-to-real gap to actual hardware.
Our proposed method yielded a $\sim 2.5\times$ increase in training sample efficiency to stable locomotion and our deployed policy was able to achieve speeds of 3.0 m/s on hardware. 

In future work we aim to push further to extend these techniques by attempting to learn even more dynamic behaviors, such as sharp $90^{\circ}$ turns.
This line of research we will also  require investigation into strategies  to stably switch between learned dynamic control policies in real time.
We also seek to develop a better understanding of what reference information is useful for maneuvers that are very different from nominal locomotion.






\bibliographystyle{IEEEtran.bst}
\bibliography{references.bib}

\end{document}